\documentclass[aps,pre,twocolumn,groupedaddress]{revtex4}
\usepackage{graphicx}
\usepackage{amsmath}
\usepackage{nicefrac}
\usepackage{color, soul}
\usepackage{algcompatible}
\usepackage{float}

\begin{document}

\title{Time Shifts to Reduce the Size of Reservoir Computers}
\author{Thomas L. Carroll}
\email{thomas.carroll@nrl.navy.mil}
\affiliation{US Naval Research Lab, Washington, DC 20375}
\author{Joseph D. Hart}
\email{joseph.hart@nrl.navy.mil}
\affiliation{US Naval Research Lab, Washington, DC 20375}

\date{\today}

\begin{abstract}
A reservoir computer is a type of dynamical system arranged to do computation. Typically, a reservoir computer is constructed by connecting a large number of nonlinear nodes in a network that includes recurrent connections. In order to achieve accurate results, the reservoir usually contains hundreds to thousands of nodes. This high dimensionality makes it difficult to analyze the reservoir computer using tools from dynamical systems theory. Additionally, the need to create and connect large numbers of nonlinear nodes makes it difficult to design and build analog reservoir computers that can be faster and consume less power than digital reservoir computers. We demonstrate here that a reservoir computer may be divided into two parts; a small set of nonlinear nodes (the reservoir), and a separate set of time-shifted reservoir output signals. The time-shifted output signals serve to increase the rank and memory of the reservoir computer, and the set of nonlinear nodes may create an embedding of the input dynamical system. We use this time-shifting technique to obtain excellent performance from an opto-electronic delay-based reservoir computer with only a small number of virtual nodes. Because only a few nonlinear nodes are required, construction of a reservoir computer becomes much easier, and delay-based reservoir computers can operate at much higher speeds.
\end{abstract}

\maketitle

\section{Introduction}
Reservoir computers were designed as recurrent neural networks that were easy to train \cite{jaeger2001, natschlaeger2002}. As in a recurrent neural network, a reservoir is typically built by connecting a large number of nonlinear nodes in a recurrent network, called the reservoir. Unlike a recurrent neural network, the connections between nodes in the reservoir are not trained; instead, a linear combination of the time series signals output by the nodes is used to fit a training signal. Because the training can be done by ridge regression, it is much faster and more stable than the training required for traditional recurrent neural networks. Additionally, the fact that the connections between nodes are fixed has led to the development of analog hardware reservoir computers, bringing the possibility of fast computation in a device with low size, weight and power.

Reservoir computers that are all or part analog include photonic systems \cite{appeltant2011,larger2012, van_der_sande2017, hart2019,chembo2019,argyris2019}, analog electronic circuits \cite{schurmann2004}, mechanical systems \cite{dion2018} and field programmable gate arrays \cite{canaday2018}. Many other examples are included in the review paper \cite{tanaka2019}, which describes hardware implementations of reservoir computers that are very fast, and yet consume little power, while being small and light. Reservoir computers have been shown to be useful for solving a number of problems, including reconstruction and prediction of chaotic attractors \cite{lu2018,zimmerman2018,antonik2018,lu2017,jaeger2004}, recognizing speech \cite{larger2017}, handwriting or other images \cite{jalavand2018}, network inference \cite{banerjee2019,banerjee2021}, or controlling robotic systems \cite{lukosevicius2012} . Reservoir computers have also been used to better understand the function of neurons in the brain \cite{stoop2013}.  Lymburn et al. \cite{lymburn2019} studied the relation between generalized synchronization and reconstruction accuracy, while Herteux and R{\"a}th examined how the symmetry of the activation function affects reservoir computer performance \cite{herteux}. 

The usual theory on why reservoir computers work is that the reservoir maps an input signal into a high dimensional space. This high dimensional system is then projected down to a lower dimensional system of interest by a linear projection. Hart et al. \cite{hart2020} showed that there is a positive probability that a reservoir computer can be an embedding of the driving system, and therefore it can predict the future of the driving system within an arbitrary tolerance, while Grigoryeva et al. \cite{grigoryeva2021} show conditions under which a reservoir computer can be in strong generalized synchronization with the driving system.

The requirement to map an input signal into a high dimensional space suggests that the reservoir must contain a large number of nodes, making analog reservoir computers more difficult to build. Time delay-based reservoir computers \cite{appeltant2011,larger2012, van_der_sande2017, hart2019,chembo2019} mitigate this by time-multiplexing a single analog nonlinearity, but this comes at the cost of a speed reduction. Recent work proposes that the recurrent network can be eliminated entirely and replaced with nonlinear combinations of the past input states \cite{gauthier2021}. We suggest that there is no particular reason that a high dimensional space is necessary to form an embedding of the input system; a smooth dynamical system of $d$ dimensions requires at most $2d$ dimensions for an embedding. It is necessary for the reservoir to be in generalized synchronization with the driving system, but it is not clear what else is required.

We demonstrate here that a reservoir computer may be divided into two parts: a small recurrent network (``the reservoir''), which we presume creates the embedding, and a larger matrix consisting of the reservoir output signals transformed in a way that increases their rank and memory. We accomplish this rank-increasing transformation using a modified version of the method developed by Del Frate et al. \cite{delfrate2021}, in which values of the reservoir nodes from previous time steps are used to create a time-shifted reservoir matrix.

First, we show via simulations of three different types of reservoir computer that the rank of the covariance matrix of the reservoir output matrix can help predict the accuracy of a reservoir computer. Then, we show, again through simulations, that the time-shifted reservoir matrix generally has a significantly improved covariance rank, memory capacity, and prediction accuracy compared with the standard reservoir matrix. We confirm that these results hold in the presence of noise by performing experimental measurements on an analog opto-electronic delay-base reservoir computer. Our results show that high accuracy can be obtained with very small reservoirs, which has important implications for the speed, size, and ease of construction of analog reservoir computers.

\section{Reservoir Computers}
\label{rescomp}
A general form for a reservoir computer can be
\begin{equation}
\label{rccomp}
\mathbf{R}\left(n+1\right)=\chi \left(\mathbf{AR}\left(n\right)+\mathbf{W}s\left(n\right)\right)
\end{equation}
 where $\chi ()$ is a nonlinear function, ${\bf R}$ is the vector of the states of the reservoir nodes, ${\bf A}$ is the adjacency matrix, $s(n)$ is the input signal and ${\bf W}$ is a vector of input signal weights. The individual components of ${\bf R}(n)$ are $r_i(n)$, where $i$ is the index of a particular node. 
 
  In the training stage, the reservoir computer is driven with the input signal $s(n)$ to produce the reservoir computer output signals $r_i(n)$. In all the examples in this paper, the input signal is normalized to have a mean of zero and a standatd deviation of 1. The reservoir output matrix $\Omega_1$ is constructed from the reservoir signals as
\begin{equation}
\label{fitmat}
\Omega_1=\left[{\begin{array}{cccccc}
{{r_{1}}\left(1\right)} & \cdots & {{r_{M}}\left(1\right)}\\
{{r_{1}}\left(2\right)} & {} & {{r_{M}}\left(2\right)}\\
\vdots & {} & \vdots\\
{{r_{1}}\left(N\right)} & \cdots & {{r_{M}}\left(N\right)}
\end{array}}\right]
\end{equation}
where the reservoir has $M$ nodes and the time series have $N$ points. The fit to the training signal is 
\begin{equation}
\label{train_fit}
{h(n)} ={\Omega_1 } {{\bf C}}
\end{equation}
where ${\bf C}$ is a vector of training coefficients. The coefficients of ${\bf C}$ are obtained by minimizing the error between $h(n)$ and a training signal $f(n)$, typically using ridge regression \cite{tikhonov} to avoid overfitting.

 
 In the testing stage, a new input signal $\tilde s\left( n \right)$ is generated from the same dynamical system that generated $s(n)$, but with different initial conditions. The corresponding test signal is $\tilde f\left( n \right)$. The input signal $\tilde s\left( n \right)$ drives the same reservoir to produce the output signals ${\tilde r_i}\left( n \right)$, which are arranged in a matrix $\tilde \Omega_1$. The testing error is
 \begin{equation}
 \label{test_err}
{\Delta_{tx}}={{\mathrm{std}\left({\tilde{f}\left(n\right)-\tilde{\Omega}_{1}{\bf {C}}}\right)}\mathord{\left/\right.\kern-\nulldelimiterspace}{\mathrm{std}\left({\tilde{f}\left(n\right)}\right)}}
\end{equation}
where the coefficient vector ${\bf C}$ was found in the training stage. 

\section{Covariance Rank, Memory Capacity, and Testing Error}
\label{rankdimerr}
 Equations (\ref{train_fit} - \ref{test_err}) show that the columns of $\Omega_1$ are used as a basis to fit the training and test signals. Principle component analysis \cite{joliffe2011} states that the eigenvectors of the covariance matrix of $\Omega$, $\Theta=\Omega_1^T\Omega_1$, form an uncorrelated basis set. The rank of the covariance matrix tells us the number of uncorrelated vectors. Therefore, we will use the rank of the covariance matrix of $\Omega_1$,
\begin{equation}
\label{rank}
\Gamma  = {\rm{rank}}\left( {\Omega_1 ^T\Omega_1 } \right)
\end{equation}
to characterize the reservoir matrix $\Omega_1$. We calculate the rank using the MATLAB rank() function, which returns the number of singular values above a certain threshold. The threshold is $\gamma_r={D_{\max }}\delta \left( {{\sigma _{\max }}} \right)$, where $D_{max}$ is the largest dimension of $\Omega_1$ and $\delta(\sigma_{max})$ is the difference between the largest singular value of $\Omega_1$ and the next largest double precision number. It was shown in  \cite{carroll2019} that larger values of the covariance rank corresponded to smaller values of the testing and training error.

Memory capacity, as defined in \cite{jaeger2002}, is considered to be an important quantity in reservoir computers. Memory capacity is a measure of how well the reservoir can reproduce previous values of the input signal.

The memory capacity as a function of delay is
\begin{equation}
\label{memdel}
{\rm {M}}{{\rm {C}}_{k}}=\frac{{\sum\limits _{n=1}^{N}{\left[{s\left({n-k}\right)-\left\langle s\right\rangle }\right]\left[{{h_{k}}\left(n\right)-\left\langle h_{k}\right\rangle }\right]}}}{{\sum\limits _{n=1}^{N}{\left[{s\left({n-k}\right)-\left\langle s\right\rangle }\right]\sum\limits _{n=1}^{N}{\left[{{h_{k}}\left(n\right)-\left\langle h_{k}\right\rangle }\right]}}}}
\end{equation}
where the $\left\langle \right\rangle$ operator indicates the mean. The signal $h_k(n)$ is the fit of the reservoir signals $r_i(n)$ to the delayed input signal $s(n-k)$. The memory capacity is
\begin{equation}
\label{memcap}
{\rm{MC}} = \sum\limits_{k = 1}^\infty  {{\rm{M}}{{\rm{C}}_k}} 
\end{equation}

Input signals such as the Lorenz $x$ signal contain correlations in time, which will cause errors in the memory calculation, so in Eq. (\ref{memdel}), $s(n)$ is a random signal uniformly distributed between -1 and +1.  There are some drawbacks to defining memory in this way; the reservoir is nonlinear, so its response will be different for different input signals, and the definition involves a fit to a signal. Table \ref{table1} below shows a situation where supplementing the reservoir signals with a memoryless nonlinearity appears to increase the memory capacity, even though the nonlinearity has no memory. Nevertheless, this memory definition is the standard definition used in the field of reservoir computing.

It is generally believed that the training and testing errors for a reservoir computer are limited at least in part by the number of nodes; more nodes yield the possibility of lower error. Here we show that it is not the number of nodes that are of primary importance but the covariance rank and possibly the memory capacity of the reservoir computer. More nodes can be better because the covariance rank is limited by the number of nodes. Memory capacity can be affected by the number of nodes, although our data suggests that covariance rank has a larger effect. It has been demonstrated that too much memory capacity can degrade the performance of a reservoir computer \cite{carroll2020b,carroll2022}.

\subsection{Types of Reservoir Computers}
In order to demonstrate the generality of the impacts of reservoir size, covariance rank and memory capacity on reservoir computing accuracy, we used three different reservoir nonlinearities with two different input signals.
\subsubsection{Tanh reservoir computer}
 A reservoir computer based on a hyperbolic tangent is used because it is common in the literature \cite{inubushi2017}. The tanh reservoir computer is described by
\begin{equation}
\label{tanhnode}
{\bf{R}}\left( {n + 1} \right) = g\tanh \left( {{\bf{AR}} + \varepsilon s\left( n \right)} \right).
\end{equation}

For each realization the parameters $g$, $\varepsilon$ and the spectral radius $\rho$ of the adjacency matrix ${\bf A}$ are each chosen from uniform random distributions between 0 and 1. Half of the entries in the adjacency matrix ${\bf A}$ are set to numbers drawn from a Gaussian random distribution, while the others are zero. The reservoir computer parameters, the adjacency matrix and the input vector are different for each realization of the reservoir computer.

Because the tanh reservoir does not contain any even nonlinearities, the set of reservoir computer output signals $r_i(n)$ were supplemented with their squares, $r_i^2(n)$, as is commonly done for this type of nonlinearity.

\subsubsection{polyODE reservoir computer}
A second reservoir computer is based on a polynomial ordinary differential equation, and will be called the polyODE reservoir computer. This reservoir computer is described by
\begin{equation}
\label{polyode}
\begin{split}
 &\frac{{d{r_i}\left( t \right)}}{{dt}}  \\
& = \alpha \left[ {{p_1}{r_i}\left( t \right) + {p_2}r_i^2\left( t \right) + {p_3}r_i^3\left( t \right) + \sum\limits_{j = 1}^M {{A_{ij}}{r_j}\left( t \right)}  + {W_i}s\left( t \right)} \right].
\end{split}
\end{equation}

The adjacency matrix was chosen in the same way as for the tanh reservoir computer. The entries in the input vector ${\bf W}$ were drawn from a uniform random distribution between -1 and 1. For each realization the parameter $p_1$ was chosen from a uniform random distribution between -5 and 0, $p_2$ was between 0 and 1, $p_3$ was between -1 and 1, the adjacency matrix spectral radius $\rho$ was between 0 and 1 and the time constant $\alpha$ was between 0 and 5. The equations were integrated with a 4th order Runge-Kutta with a time step of 1.

\subsubsection{Opto-electronic delay-based reservoir computer}
The third reservoir computer is a simulation of the opto-electronic delay-based reservoir computer of Ref. \cite{larger2012}. In this reservoir computer multiple virtual nodes were created by time-multiplexing a driven opto-electronic oscillator.  The virtual nodes were coupled because the delay loop contained a low pass filter, so each virtual node was coupled to the preceding nodes \cite{hart2019}.

The opto-electronic reservoir computer can be modeled by
\begin{equation}
\label{laseq}
{T_{L}}\dot{\nu}(t)=-\nu\left(t\right) + \beta\sin^2{\left({\nu\left({t-{\tau_{D}}}\right)+\phi+\rho W\left(t\right)s_{in}\left(t\right)}\right)}.
\end{equation}

As with the other reservoir computers, the time step for this model was fixed at $t_S=1$. The time per node was $\theta=50$, which was chosen to give a large number of time steps per node. The total delay time was $\tau_D=M_1 \times \theta$, where $M_1$ was the number of virtual nodes. The input mask $W(t)$ is a piecewise constant function of time defined as $W(t)=W_k$ where $k=(t {\rm mod} \tau_D)/\theta$, where the elements of the length $M_1$ vector $W_k$ were randomly set to either 1 or -1. Note that $W$ is periodic in $\tau_D$. 

Equation \ref{laseq} was integrated with a 4th order Runge-Kutta routine. The 4th order routine evaluates the function at the beginning and end of the integration interval. For the time $t$ and $t+1$, the input signals were
\begin{equation}
\label{sin_eq}
{s_{in}}\left( t \right) = s\left( {\left\lfloor {\frac{t}{{{\tau _D}}}} \right\rfloor } \right)\quad {s_{in}}\left( {t + 1} \right) = s\left( {\left\lfloor {\frac{{t + 1}}{{{\tau _D}}}} \right\rfloor } \right)
\end{equation}
where $\left\lfloor  \right\rfloor$ is the floor operator. 

After numerical integration, the time series $\nu(t)$ was reshaped into reservoir computer form as
\begin{equation}
\label{rcmat}
\Omega_1=\left[\begin{array}{cccc}
\nu\left(\theta\right) & \nu\left(2\theta\right) & \ldots & \nu\left(M\theta\right)\\
\nu\left(\theta+\tau_{D}\right) &  &  & \vdots\\
\vdots &  &  & \vdots\\
\nu\left(\theta+N\tau_{D}\right) & \ldots & \ldots & \nu\left(M\theta+N\tau_{D}\right)
\end{array}\right]
\end{equation}

For each realization, the parameter values were drawn from uniform random distributions. For $T_L$ the limits were 0 to 300, for $\rho$ and $\beta$ the limits were 0 to 1, and for $\phi$ the limits were 0 to $\pi$.

\subsection{Reservoir observer task}
\label{observer}
We qualify these reservoir computers by their ability to perform the ``observer task'' of performing model-free inference of unmeasured variables  \cite{lu2017}. To demonstrate generality, we test the reservoir computers' ability to perform the observer task on both the Lorenz system and the R{\"o}ssler system; that is, we ask the reservoir computer to infer the $z$ variable given only the $x$ variable for a given system.

The input signals for the simulations were taken from the Lorenz or R{\"o}ssler chaotic systems. The Lorenz system is described by \cite{lorenz1963}
\begin{equation}
\label{loreq}
\begin{array}{l}
\frac{{dx}}{{dt}} =T_l \left( {p_1}\left( {y - x} \right) \right)\\
\frac{{dy}}{{dt}} = T_l \left( x\left( {{p_2} - z} \right) - y\right)\\
\frac{{dy}}{{dt}} = T_l \left( xy - {p_3}z\right)
\end{array}
\end{equation}
with $p_1=10$, $p_2=28$, $p_3=8/3$ and $T_l=0.1$. The Lorenz equations were numerically integrated with a time step of 1. The input to the reservoir computer was $x$, and the reservoir computer was trained to output $z$. 

The R{\"o}ssler system is described by \cite{rossler1976}
\begin{equation}
\label{rosseq}
\begin{array}{*{20}{l}}
{\frac{{dx}}{{dt}} = T_r \left( - y - {p_1}z \right)}\\
{\frac{{dy}}{{dt}} = T_r \left(x + {p_2}y \right)}\\
{\frac{{dz}}{{dt}} = T_r \left( {p_3} + z\left( {x - {p_4} } \right) \right)}
\end{array}
\end{equation}

The R{\"o}ssler equations were numerically integrated  with a time step of 1, $T_r=0.65$ and parameters $p_1=1$, $p_2=0.2$, $p_3=0.2$, $p_4=5.7$.  As with the Lorenz system, the input to the reservoir computer was $x$, and the reservoir computer was trained to output $z$. 

\subsubsection{Results}

\begin{figure*}
\centering
\includegraphics[scale=0.8]{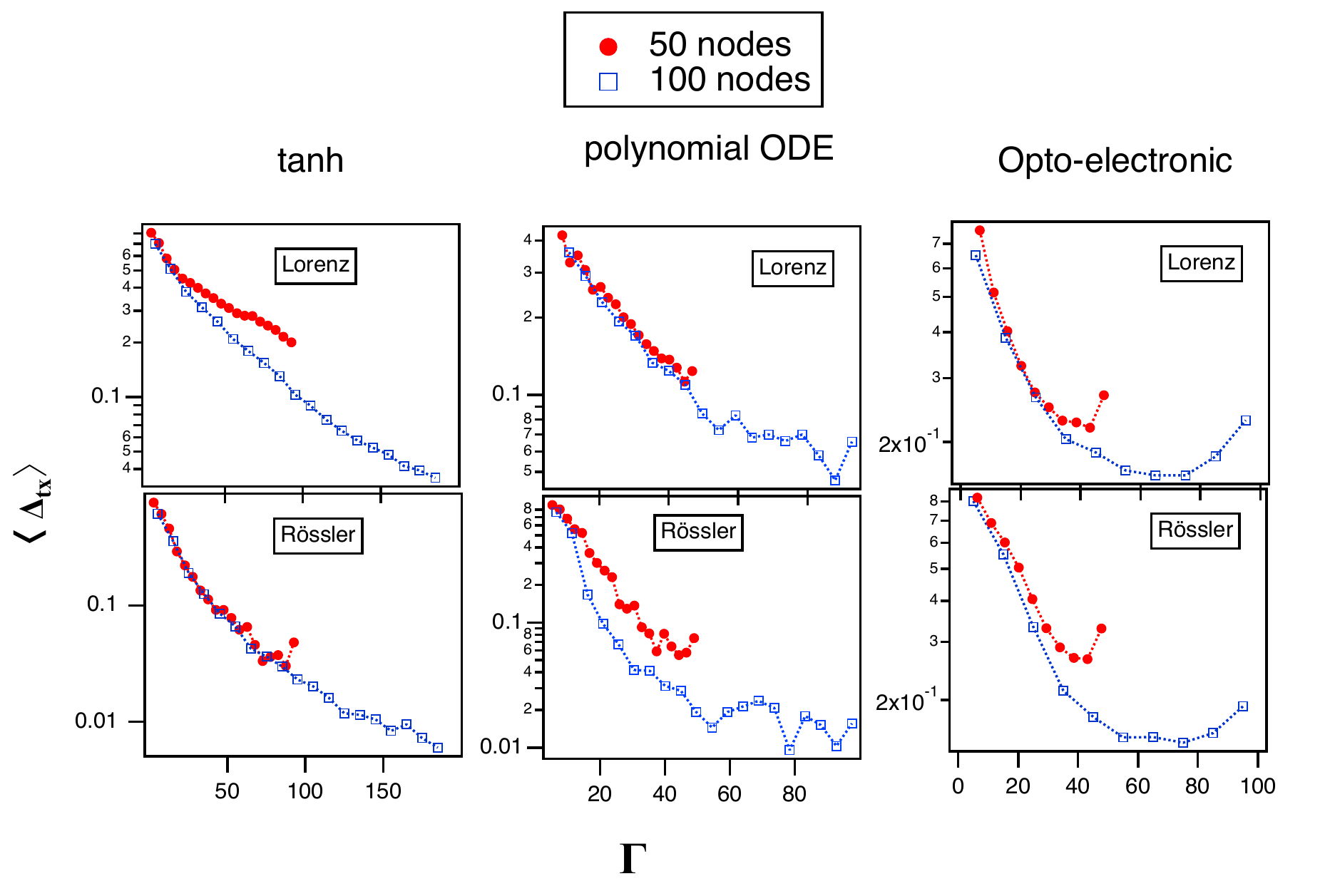}
\caption{\label{errrank} Mean testing error $\left <\Delta_{tx} \right >$ as a function of covariance rank $\Gamma$ for Lorenz or R{\" o}ssler $x$ signals driving either a tanh, a polynomial ODE or a opto-electronic reservoir computer. The reservoir computers were trained on the $z$ signals from the corresponding input systems. Reservoir computers containing 50 or 100 nodes were used. The tanh data has twice the rank because the set of reservoir computer signals $r_i(t)$ were supplemented with their squares $r_i^2(t)$.}
\end{figure*}

Figure \ref{errrank} shows the mean testing error $\Delta_{tx}$ as a function of covariance rank $\Gamma$ for six combinations of input signal and reservoir computer.

In Fig. \ref{errrank} the mean testing error depends on the covariance rank $\Gamma$, not on the number of nodes in the reservoir computer. The tanh data has twice the rank because the set of reservoir computer signals $r_i(t)$ were supplemented with their squares $r_i^2(t)$. Using more nodes allows the covariance rank to be larger, since it is limited by the size of the reservoir. For a particular rank there is a spread in testing error because there are other factors that affect the error, such as the reservoir Lyapunov exponents \cite{carroll2020b} or the memory \cite{carroll2022}, but the main effect of the number of nodes is to increase the possible rank. The point of Fig. \ref{errrank} is that the curve of testing error versus rank does not depend on the number of nodes in the reservoir computer, but rather on the covariance rank.

Figure \ref{errrank} does show that for the highest ranks in the laser system the testing error actually increases. The laser nonlinearity is bounded, so for some parameter combinations it is possible the laser equation has one or more positive Lyapunov exponents. Positive Lyapunov exponents would lead to complex chaotic signals that had high rank, but they would also cause the breakdown of generalized synchronization, so the testing error would be large.

\begin{figure*}
\centering
\includegraphics[scale=0.8]{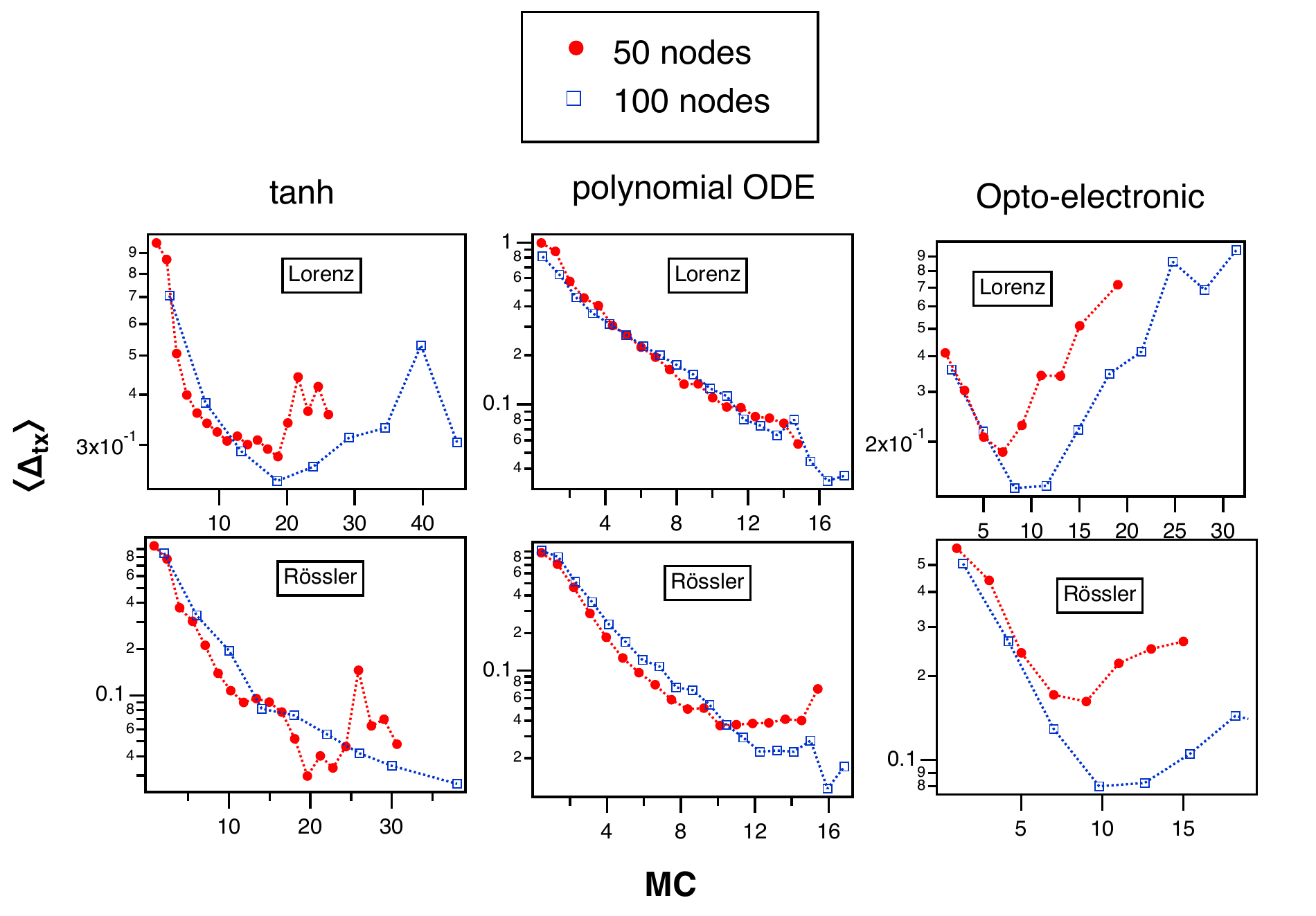}
\caption{\label{errmem} Mean testing error $\left <\Delta_{tx} \right >$ as a function of memory capacity MC for Lorenz or R{\" o}ssler $x$ signals driving either a tanh, a polynomial ODE or a opto-electronic reservoir computer. The reservoir computers were trained on the $z$ signals from the corresponding input systems. Reservoir computers containing 50 or 100 nodes were used. }
\end{figure*}

It is generally accepted that the larger memory capacity for a reservoir computer leads to lower testing errors, although \cite{carroll2022} pointed out that the memory capacity must be matched to the problem at hand, so memory capacity in some cases can be too large. We may ask if larger reservoir computers increase the memory capacity; Fig. \ref{errmem} and Table \ref{table1} show that the memory capacity increases as the reservoir computer becomes larger, but the increase scales slower than linear with the number of nodes.

Table \ref{table1} shows a shortcoming in the standard definition of memory capacity in Eqs. (\ref{memdel}-\ref{memcap}). Table \ref{table1} shows the measured memory capacity for the tanh reservoir computers with 50 or 100 nodes. The training signal may be fit using only the reservoir signals $r_i(t)$ or supplementing these signals with $r_i^2(t)$. Table \ref{table1} shows that the increase in calculated memory capacity when supplementing the reservoir signals with their squares is comparable to the increase in memory capacity when doubling the number of nodes. The operation of squaring the reservoir signals is a memoryless nonlinearity, however, so it should not change the actual memory of the reservoir computer. Adding the squared signals does improve the reservoir computer fit to the training signal, causing an apparent increase in memory capacity when there is no actual increase.

\begin{table}
\caption{ Mean memory capacity when a uniformly distributed random signal drove a tanh reservoir computer, where either $r_i(t)$ or $r_i(t)$ and $r_i^2(t)$ are used to fit the training signal}
\label{table1}
\begin{tabular}{ | c | c | c | c |}
\hline
   50 nodes $r_i(t)$ & 50 nodes $r_i(t)$, $r_i^2(t)$ & 100 nodes $r_i(t)$ & 100 nodes $r_i(t)$, $r_i^2(t)$\\
\hline
 6.2 & 7.0 & 7.5 & 9.2 \\
\hline
\end{tabular}
\end{table}

Figure \ref{errmem} does show that, as with the covariance rank, the curve of testing error versus memory capacity mostly follows the same curve independent of the number of nodes. The curves do diverge in the tanh and opto-electronic reservoirs when the testing error actually increases with memory capacity.  Memory capacity is not as simple as covariance rank because effects that increase memory capacity, such as increased Lyapunov exponents, can also increase testing error \cite{carroll2020b}. 

Figures \ref{errrank} and \ref{errmem} suggest that in order to reduce the testing error, we would like to increase the covariance rank and find the optimum memory capacity. The following section (\ref{simulations}) shows that increasing the rank or memory does not necessarily require increasing the size of the reservoir; we find that using a matrix of time-delayed signals from the reservoir can give us the necessary rank or memory. We find that the reservoir itself requires only a small number of nodes. Creating and connecting nonlinear nodes in analog hardware can be difficult, so using small reservoir computers should make them much easier to build.

\section{Time-Shifted Reservoir Computers}
\label{simulations}
Section \ref{rankdimerr} showed that increased covariance rank leads to better reservoir computer performance, regardless of the number of nodes. In a sense, this is what was done when augmenting the tanh reservoir with the $r_i^2$ values. One way to optimize reservoir computer performance therefore is to maximize the rank for a given number of nodes. One way to do this is to oversample the reservoir; this has been shown to improve performance for small oversampling factors such that the number of ``effective nodes'' is increased by a factor of 2-3 \cite{takano2018,harkhoe2020,sunada2021}. Jaurigue et al. \cite{jaurigue2021} showed that augmenting the reservoir matrix with a delayed version of the input signal can also improve the quality of time series prediction.

Another way to increase the covariance rank is based on the time shifting method of Del Frate et al. \cite{delfrate2021}; this technique can result in a very large number of ``effective nodes''. In that paper, the authors applied random time shifts to the set of reservoir node states and showed that this approach could reduce the testing error. Their goal was to improve the reservoir computer performance without optimizing all parameters. A reduced version of this method using a single time shift  has been shown to improve the performance of a photonic analog reservoir computer \cite{harkhoe2020}.

We have simplified the method of \cite{delfrate2021} by using an ordered set of time shifts instead of random shifts. In this case our goal was to see if we could achieve a small testing error with only a small number of nodes, so we simulated a reservoir computer with $M_1$ nodes and used the output signals to create a matrix of $M_2$ signals, where $M_2 \ge M_1$. 

We decided to perform these simulations for the opto-electronic reservoir computer of Eq. (\ref{laseq}) because a reduction in $M_1$ results in a proportional increase in the computation speed of delay-based reservoir computers. Further, we are able to confirm these simulations with a laboratory experiment (see section \ref{lasexp}). The parameters for these simulations were $T_L=200$, $\beta=0.5$, $\rho=1$, $\theta=50$ and $\phi=\pi/4$. 

The matrix of signals from the reservoir computer was $\Omega_1$ as in Eq. (\ref{rcmat}).  The time-shifted reservoir was

\begin{equation}
\label{rcmatshift}
\Omega_{2}\left(i,j\right)=\nu \left(k\theta+\left(i-1\right)\tau_{D}-\tau_{j}\right)
\end{equation}

where $k={\rm mod}(j,M_1)$, $i=1,2,\ldots,N$ and $j=1,2,\ldots,M_2$

The maximum time shift value was $\tau_{max}$, while the individual time shifts were $\tau_j=j \times \tau_{max}/M_2$. When the time shift $\tau_j$ was not an integer multiple of $\theta$, linear interpolation was performed between the two nearest $\theta$-sampled values of the opto-electronic reservoir computer output signal $\nu$ to get the shifted values. In \cite{delfrate2021} the delay where the autocorrelation of the input signal was reduced to half its peak value was used to set $\tau_{max}$; we found that because the R{\"o}ssler system had a very slowly decaying autocorrelation, that method gave a value of $\tau_{max}$ that was too large. Instead we simulated the testing error as $\tau_{max}$ was varied with $M_1=M_2=100$ nodes and picked a value that gave a small testing error. 

\subsection{Vary Number of Nodes}
\begin{figure}
\centering
\includegraphics[scale=0.8]{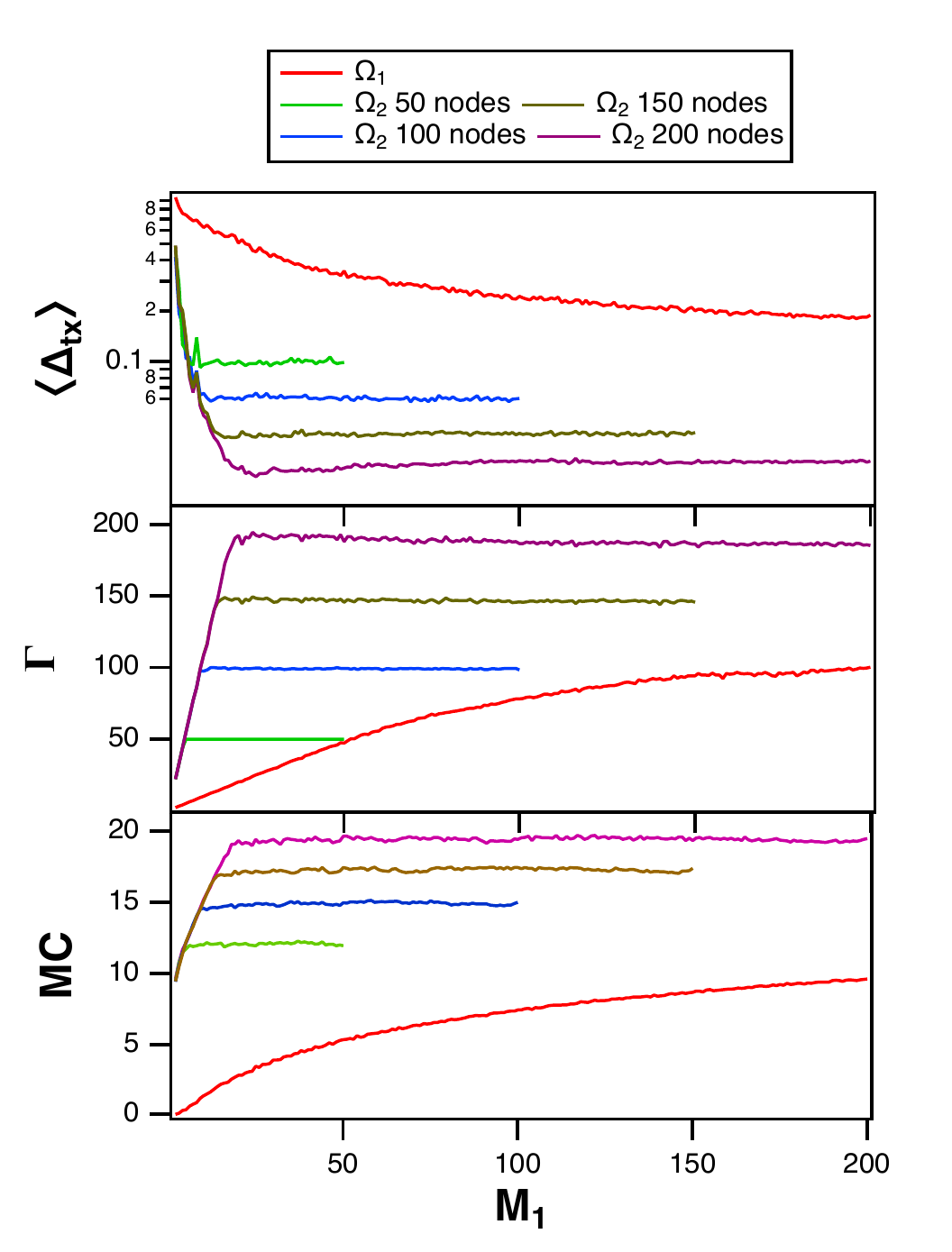} 
  \caption{ \label{laserlorshifterr} The top plot is the mean testing error for the opto-electronic reservoir computer when the input signal is the Lorenz $x$ signal and the training signal is the $z$ signal. The trace labeled $\Omega_1$ is the error for the original reservoir computer with $M_1$ nodes. The traces labeled as $\Omega_2$ are for the time-shifted matrix with different fixed values of $M_2$. The middle plot shows the covariance rank $\Gamma$ while the bottom plot shows the memory capacity MC.}
  \end{figure}

To find out how many actual nodes are needed for a reservoir computer, the laser delay reservoir computer was simulated for a variety of reservoir sizes  $M_1=2,3, \ldots M_2$ and with a fixed size $M_2$ for the time-shifted matrix.  The first task was the Lorenz observer task as described in Sec. \ref{observer}, in which case the maximum delay value $\tau_{max}$ was 10$\tau_D$, which was equivalent to 10 Lorenz time steps. The parameters for the Lorenz system were the same as in Eq. (\ref{loreq}). Figure \ref{laserlorshifterr} shows the training error, covariance rank and memory capacity based on either $\Omega_1$ or $\Omega_2$ as $M_1$ varies.  The elements of the input vector ${\bf W}$ were randomly assigned to -1 or 1. Because the training and testing error can depend on the input vector, 20 different realizations were performed, each with a different input mask, and all reported results are the mean values over these 20 realizations.

  In Fig. \ref{laserlorshifterr}, the covariance rank for the reservoir output matrix $\Omega_1$ increases with the number of nodes, but the increase slows as the number of nodes increases. Except for the case where the time-shifted reservoir matrix contains $M_2=50$ nodes, the covariance rank of the shifted reservoir matrix is always higher than the covariance rank of the unshifted matrix $\Omega_1$. The rank of the time-shifted reservoir saturates when $M_1$ is approximately equal to $M_2/10$. The time-shifted reservoir matrices also have a larger memory capacity than the unshifted reservoir matrix. Simply adding a delayed input signal with a delay of $\tau_{max}$ can increase the memory capacity, but Fig. \ref{laserlorshifterr} shows that adding more nodes also increases the memory capacity. Whether this is an actual increase of memory or an artifact of the better fit to the training signal used in the definition of memory capacity is not possible to say.
  
 Figure \ref{laserlorshifterr} also shows that the testing error for the time-shifted reservoir computer is always lower than the error for the unshifted version. This agrees with \cite{delfrate2021}, where the time-shifted reservoir with random time shifts was used to decrease the testing error in a large (N=100) reservoir computer.    
  
  What Fig. \ref{laserlorshifterr} shows is that the reservoir itself only needs a few nodes to give a small training error. The set of signals in the matrix $\Omega$ still needs a large covariance rank, but this rank may be accomplished by creating a matrix with time shifts. There appear to be two phenomena at work here; the networked nonlinear nodes in the reservoir produce enough information to re-create an approximation of the Lorenz $z$ signal, possibly by creating an embedding of the Lorenz system, while the time-shifted reservoir increases the covariance rank to improve the approximation.
  
 The time-shifting technique was also tested on a simulated opto-electronic reservoir computer driven by the R{\" o}ssler $x$ signal and trained on the R{\" o}ssler $z$ signal. In this case the maximum time shift $\tau_{max}$ was 20$\tau_D$. The parameters for the R{\" o}ssler system were the same as those used in Eq. (\ref{rosseq}).
   \begin{figure}
\centering
\includegraphics[scale=0.8]{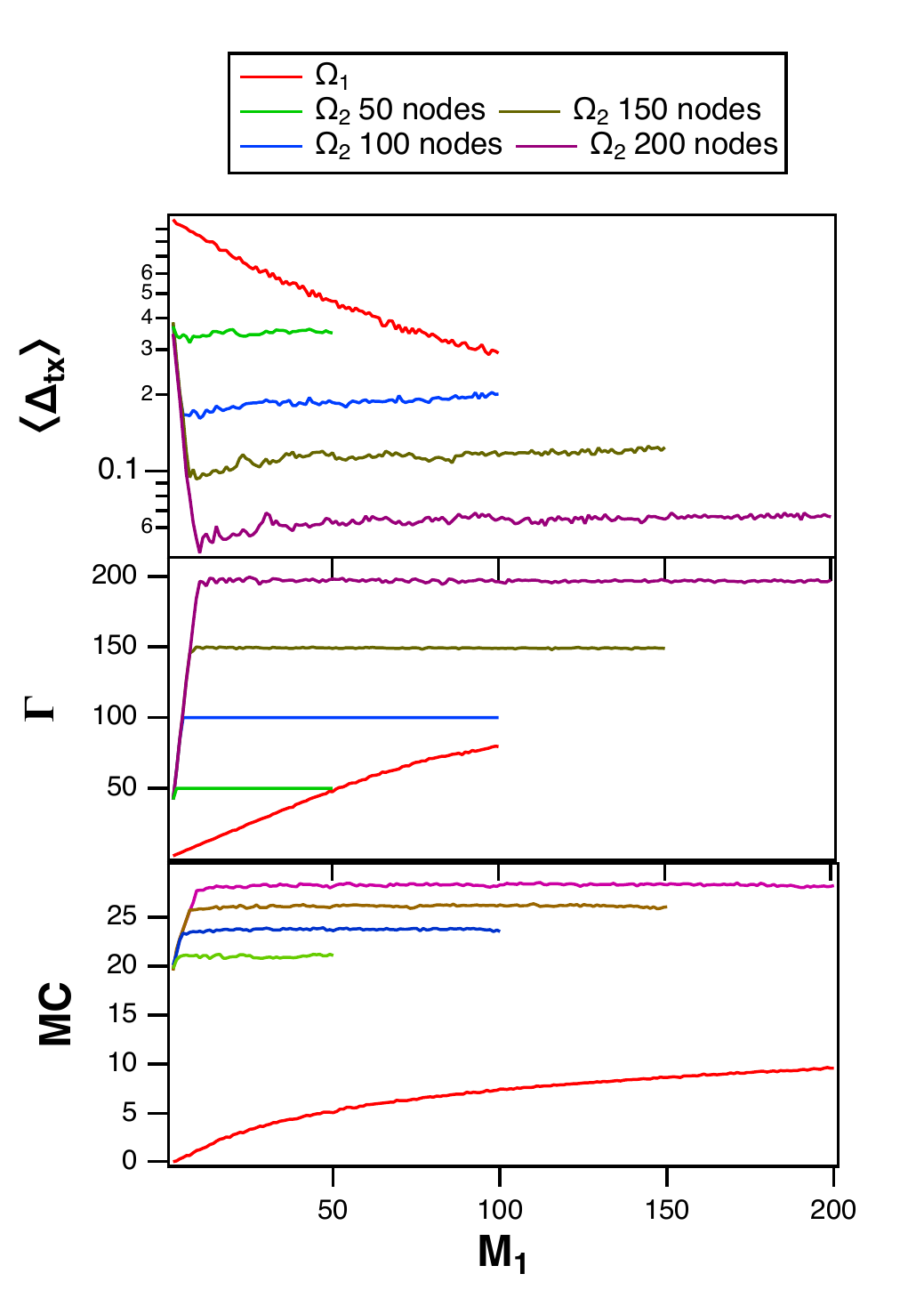} 
  \caption{ \label{laserrossshifterr} The top plot is the mean training error for the opto-electronic reservoir computer when the input signal is the R{\" o}ssler $x$ signal and the training signal is the $z$ signal. The trace labeled $\Omega_1$ is the error for the original reservoir computer with $M_1$ nodes. The traces labeled as $\Omega_2$ are for the time-shifted matrix with different fixed values of $M_2$. The bottom plot shows the covariance rank $\Gamma$. }
  \end{figure}

Figure \ref{laserrossshifterr} shows that the covariance rank for the time-shifted reservoir matrix saturates when $M_1 \approx M_2/20$. For the Lorenz system, this number was $M_2/10$. As with the Lorenz system, adding time-shifted signals to the opto-electronic reservoir computer driven by a R{\" o}ssler signal decreases the testing error.

\section{Experiment}
\label{lasexp}
\begin{figure}
\centering
\includegraphics[scale=0.8]{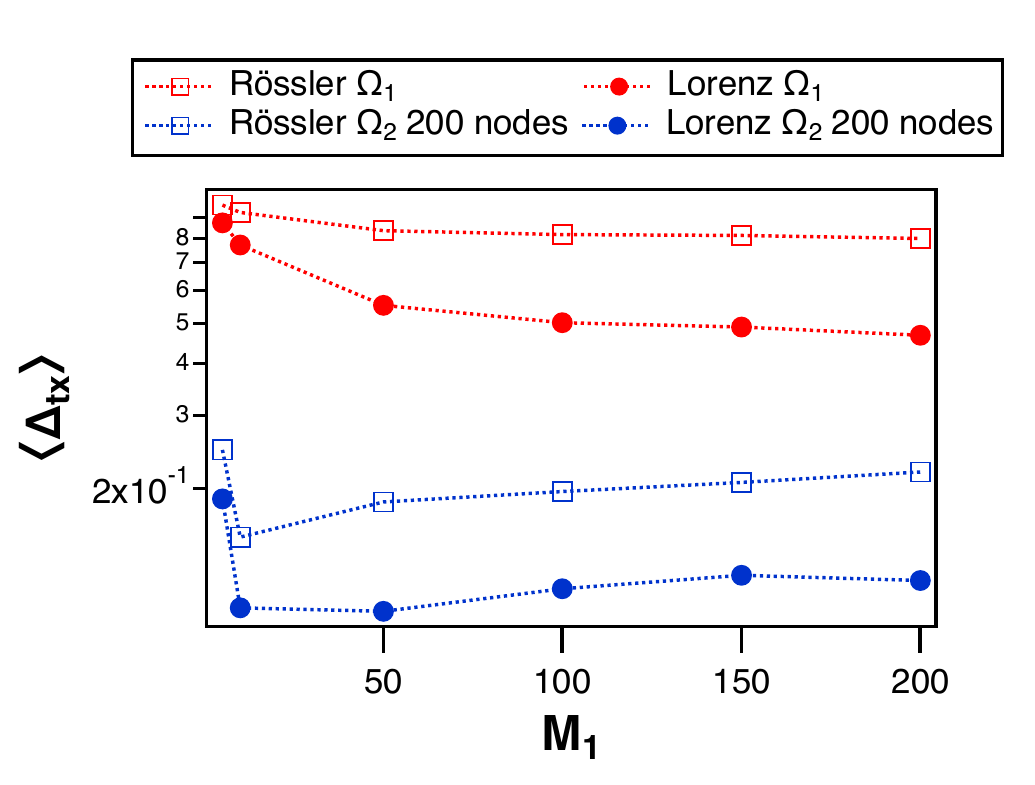} 
  \caption{ \label{experr} Mean testing errors $\left< \Delta_{tx} \right>$ as a function of the number of nodes $M_1$ in the non-time-shifted reservoir computer. The red symbols are for the reservoir computer without time shifts, while the blue symbols are for fits using the time-shifted matrix $\Omega_2$ with 200 nodes. These plots follow the same trends as the simulations in Figs. \ref{laserlorshifterr} and \ref{laserrossshifterr}.}
  \end{figure}

We performed an experiment to confirm that the simulations on the opto-electronic reservoir computer from Section \ref{simulations} hold in the real world, where noise can degrade reservoir computer performance \cite{soriano2013, semenova2019}. The experiment was similar to that described in \cite{larger2012,hart2019}; the details are given in the supplementary material \cite{}. Based on the model Eq. (\ref{laseq}), the parameters were the same as in the simulations: $\beta=0.5$, $\rho=1$, $\phi=\pi/4$ and $T_L=4 \theta $. The experiment was run with $M_1=\tau_D/\theta$ set to 5, 10, 50, 100, 150 or 200 virtual nodes. The number of time-shifted signals used in the time-shifted matrix $\Omega_2$ was fixed at $M_2=200$. As in the simulations, the elements of the input mask ${\bf W}$ were randomly set to either $\pm 1$. For each value of $M_1$ the experiment was repeated with 20 different input masks, and the testing errors and ranks shown are the mean error and rank for all 20 experiments.  The input signal $s_{in}$ was set to the Lorenz or R{\" o}ssler $x$ variable for $8000 \times \tau_D$, then set to zero for $100 \times \tau_D$ to allow the experiment to reset, and then again set to the $x$ variable for $4000 \times \tau_D$. The opto-electronic reservoir response signal $\nu(t)$ was sampled at intervals of $\theta$ and arranged in a reservoir matrix ($\Omega_1$) or time-shifted matrix ($\Omega_2$) as in Eqs. (\ref{rcmat}) and (\ref{rcmatshift}). After removing an initial transient of $200 \times \tau_D$ reservoir time steps the next $7800 \times \tau_D$ time steps were used to train the experiment. Because the 8000 point initial signal was followed by 100 points set to zero, the 3800 time series points from 8300 to the end were used for training the experiment. The number of signals in the time-shifted matrix was $M_2=200$. The mean testing errors for the experiment are plotted in Fig. \ref{experr}.

As in the simulations, using a matrix of time-shifted reservoir states $\Omega_2$ significantly decreases the testing error for both the Lorenz and R{\"o}ssler observer tasks. The testing error for the time-shifted matrix bottoms out for $M_1=5$ for these experiments, meaning that one may obtain small testing errors for a small reservoir size. In delay-based reservoir computers like our opto-electronic one, the computation rate is inversely proportional to the number of nodes. By using our time-shifting technique with $M_1=5$ reservoir nodes and $M_2=200$, we are able to increase the computation rate by a factor of 40 while decreasing the training error by more than a factor of 2 compared to traditional reservoir computing with $\Omega_1=200.$

\section{Conclusions}
In this work we showed that excellent reservoir computing performance can be obtained by taking the signals from a small reservoir and creating a matrix of time-shifted signals. We demonstrated this technique in simulations of three different types of reservoir computer and in an opto-electronic analog reservoir computer experiment. We ascribed this improvement in testing error to an increase in the covariance rank and memory capacity of the time-shifted reservoir matrix compared to the reservoir matrix without time shifts. Depending on the relative sizes of the reservoir computer and the time-shifted matrix, the improvement in testing error was in the range of 10 to 20. An immediate benefit of this technique is that the same or better testing error can be accomplished with a much smaller reservoir computer, an important consideration if creating and coupling a large number of nodes is difficult or expensive. In addition, choosing an optimal reservoir network should be much easier when only a few nodes are required.

This work also illuminates the components that are necessary to create a reservoir computer. The network of nonlinear nodes creates nonlinearity and memory, while the time-shifted matrix increases the rank and memory. It is known from work on nonlinear series approximations \cite{boyd1985,orcioni2014,gauthier2021} that memory in the form of delays may be combined with memoryless nonlinearities to approximate signals, so it is not surprising in hindsight that the nonlinearity and memory may be distributed differently. 

 \section{Data Availability}
The data that support the findings of this study are available on request from the corresponding author. The data are not publicly available because they have not been approved for public release.

\section{Acknowledgements}
This work was supported by the Naval Research Laboratory's Basic Research Program and the Department of Defense Applied Research for Advancement of S\&T Priorities Neuropipe program..

\section{Supplemental Materials}

The experimental reservoir computer (RC) is based on an opto-electronic oscillator with time-delayed feedback, similar to the ones used in Refs. \cite{paquot2012optoelectronic,larger2012,antonik2017brain,dai2021classification}. A traditional opto-electronic oscillator consists of an opto-electronic nonlinearity, an optical delay line, and photoreceiver, the output of which is fed back to the electrical input of the opto-electronic nonlinearity \cite{chembo2019optoelectronic}. Typically, the nonlinearity is an intensity modulator and the photoreceiver also acts as an analog lowpass or bandpass filter. Properly optimized opto-electronic reservoir computers have been shown to be capable of extremely high data processing rates \cite{larger2017}. An extensive review of opto-electronic reservoir computers can be found in Ref. \cite{chembo2020machine}.

In this work, we use a hybrid analog-digital opto-electronic oscillator similar to those used in Refs. \cite{antonik2017brain,murphy2010complex,hart2019laminar}. The main difference from the traditional opto-electronic oscillators is that the delay line and/or the filtering are implemented in digital electronics. This setup provides greater flexibility and control over the delay and filter parameters, and has even allowed for the implementation of arbitrary networks of coupled maps to study basic network science \cite{hart2017} and reservoir computing \cite{hart2019,soriano2019optoelectronic}; however, it does cap the maximum speed of the device to be somewhat slower than fully analog opto-electronic oscillators. 

\begin{figure*}
    \centering
    \includegraphics[width=\textwidth]{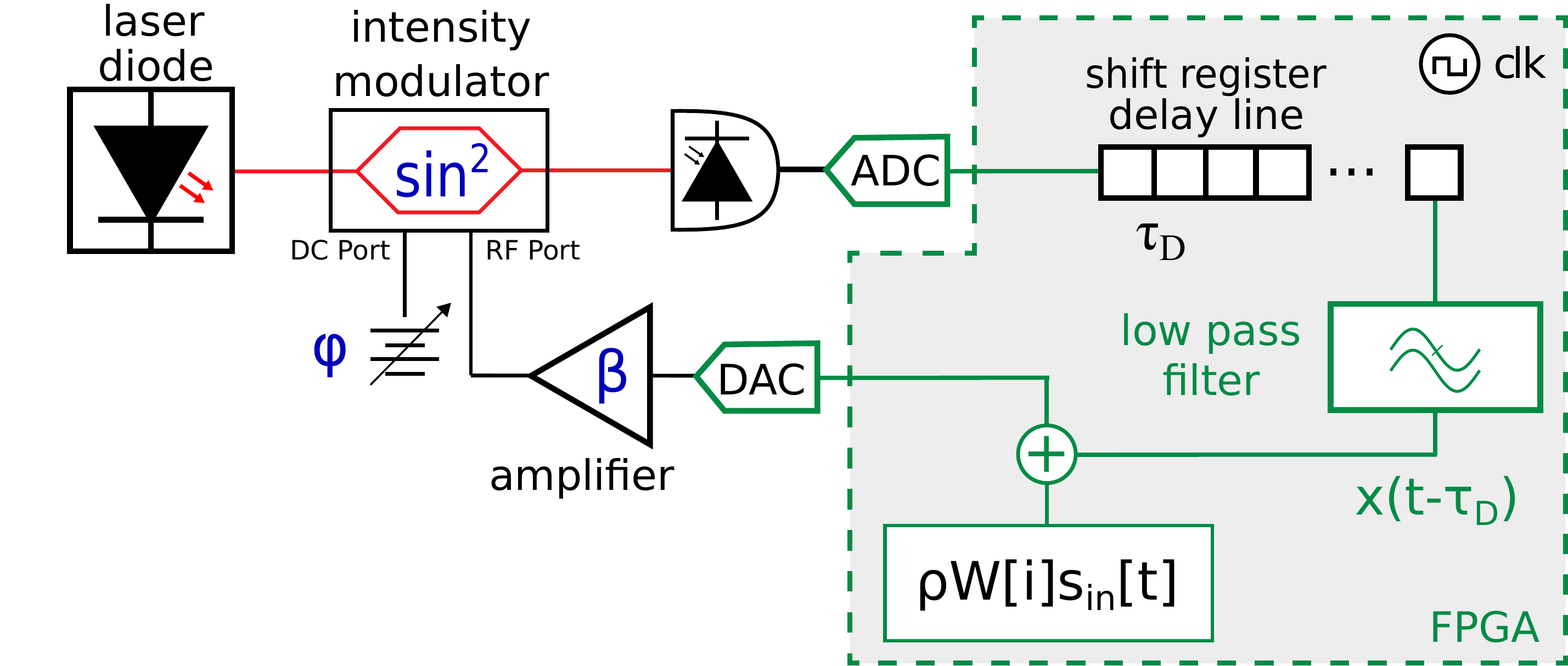}
    \caption{Schematic of the opto-electronic reservoir computer used here. The intensity modulator implements an opto-electronic sinusoidal nonlinearity. The operations in the shaded green box occur digitally inside the FPGA. The FPGA clock rate $F_{clk}=2$ MHz, which is sufficiently fast relative to $\theta=10$ $\mu$s that the discretization in time is negligible.}
    \label{fig:OEO}
\end{figure*}

A schematic of the opto-electronic oscillator used here is shown in Fig. \ref{fig:OEO}. A fiber-coupled CW laser emits light of constant intensity at 1550 nm. That light passes through a fiber-coupled intensity modulator and converted to an electrical signal by a photoreceiver (Terahertz Technologies TIA-525). The electrical signal is read into a field-programmable gate array (FPGA) by an analog to digital converter (ADC). An digital electronic delay and digital lowpass filtering are implemented on the FPGA. Additionally, the input (drive) signal is added electronically in the FPGA before it is output via a digital to analog converter (DAC). The ADC (AD9254) and DAC (TI DAC 5672) used here have 14-bits of resolution. The DAC output is amplified and used to drive the RF-port of the intensity modulator. 

The parameter $\beta$ represents the round-trip gain and is controlled by the amplifier gain on the DAC output. A variable voltage power supply drives the DC-port of the intensity modulator and is used to control the bias phase $\phi$.

The digital filter used here is a single-pole lowpass infinite impulse response (IIR) filter. The filter sampling rate $F_{clk}=2$ MHz and the filter time constant $T_L$=40 $\mu$s so that the digital filter is an excellent approximation of an analog single-pole lowpass filter. The node time $\theta$  (i.e., the time over which the input is held constant) is chosen to be 10 $\mu$s so that $\tau_L/\theta=4$. 

We choose to operate our opto-electronic oscillator at such a slow rate (relative to the state of the art) in order to reduce the cost and complexity of the electronics required. This is sufficient for our proof of principle experiments that show that adding ordered time shifts to a reservoir computer output can dramatically improve performance of a photonic reservoir computer, even in the presence of noise and slight parameter drifts inherent in any experimental setup. This same time shifting technique can be applied directly to higher rate photonic reservoir computers to either reduce the number of nodes or increase the accuracy.

\end{document}